\documentclass[9pt]{article}
\usepackage{spconf,amsmath,graphicx,hyperref}
\usepackage{tabularx}   
\usepackage{booktabs}   
\usepackage{subcaption} 
\usepackage{adjustbox}   
\usepackage{makecell}    
\usepackage{xcolor}
\usepackage{boldline}

\title{Automated Dysphagia Screening Using Noninvasive Neck Acoustic Sensing}
\name{
\parbox{\linewidth}{
\centering
Jade Chng$^{1,2,}$\sthanks{These authors contributed equally to this work.},
Rong Xing$^{1,}$\footnotemark[1],
Yunfei Luo$^{3,}$\footnotemark[1],
\\
Kristen Linnemeyer-Risser$^{\;4}$,
Tauhidur Rahman$^{1,3}$, 
Andrew Yousef$^{\;4,}$\sthanks{Equal Senior Authors},
Philip A Weissbrod$^{\;4,}$\footnotemark[2]
}
}

\address{
$^{1}$ Jacobs School of Engineering, University of California San Diego \\
$^{2}$ Department of Biomedical Engineering, Duke University \\
$^{3}$ Halıcıoğlu Data Science Institute, University of California San Diego \\
$^{4}$ Department of Otolaryngology-Head and Neck Surgery, University of California San Diego
}
\begin{document}
%
\maketitle
\begin{abstract}
Pharyngeal health plays a vital role in essential human functions such as breathing, swallowing, and vocalization. Early detection of swallowing abnormalities, also known as dysphagia, is crucial for timely intervention. However, current diagnostic methods often rely on radiographic imaging or invasive procedures. 
In this study, we propose an automated framework for detecting dysphagia using portable and noninvasive acoustic sensing coupled with applied machine learning. By capturing subtle acoustic signals from the neck during swallowing tasks, we aim to identify patterns associated with abnormal physiological conditions. 
Our approach achieves promising test-time abnormality detection performance, with an AUC-ROC of 0.904 under 5 independent train–test splits. 
This work demonstrates the feasibility of using noninvasive acoustic sensing as a practical and scalable tool for pharyngeal health monitoring.
\end{abstract}

\begin{figure*}[ht]
\vspace{-8mm}
\centerline{\includegraphics[width=0.97\textwidth]{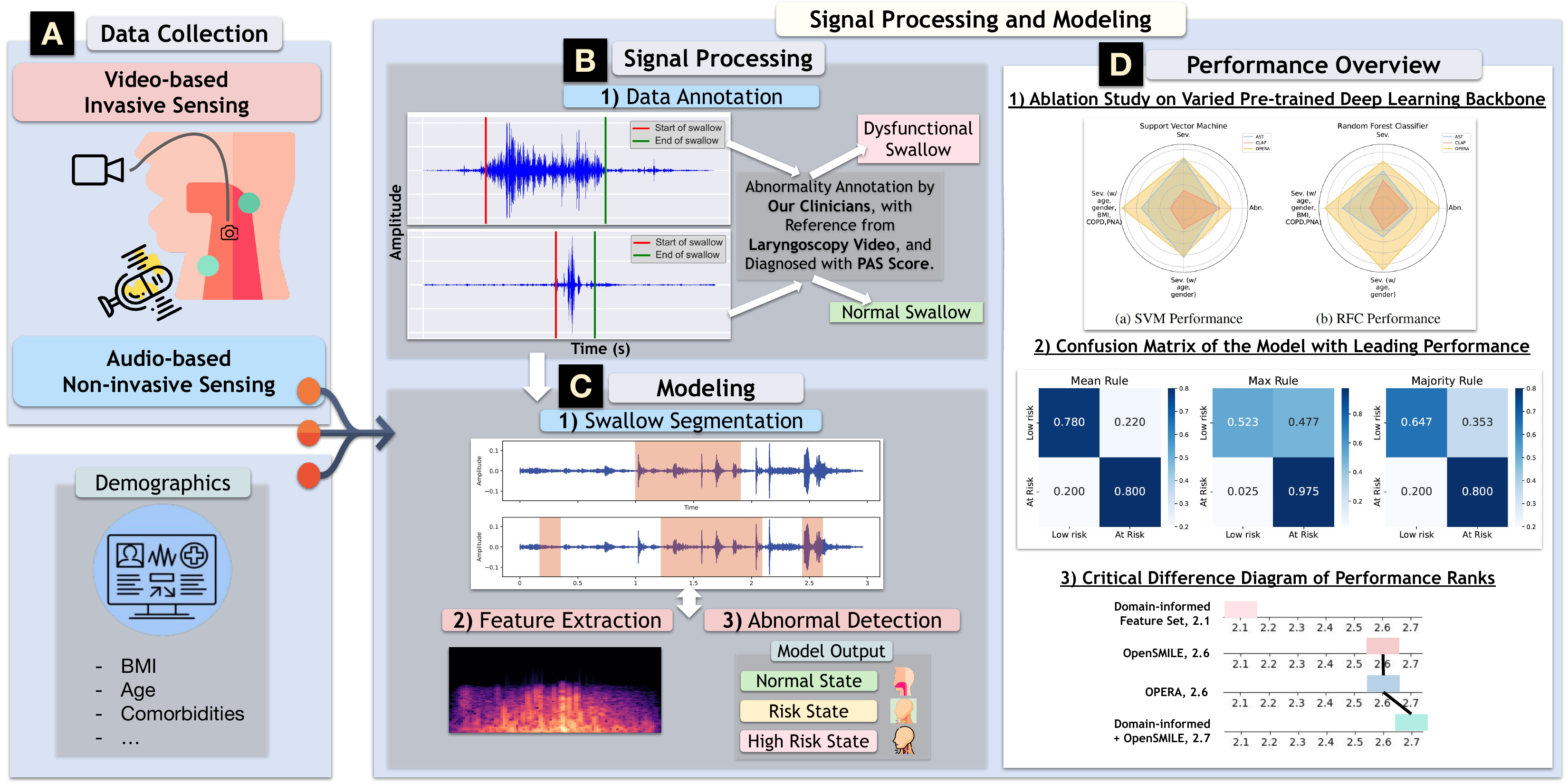}}
\vspace{-3mm}
\caption{Overview of our proposed automated system. (A) Data collected from participants during this study. (B) Demonstration of our data annotation process. (C) Modeling procedure to explore the relationship between acoustic signal and the swallow abnormalities. (D) Presentation of the empirical results. }
\label{fig:overview}
\vspace{-4mm}
\end{figure*}

\section{Introduction}
\label{sec:intro}
\vspace{-1mm}
Dysphagia, or difficulty swallowing, continues to pose a significant public health concern, affecting 10-20\% of adults over the age of 50 and up to two-thirds of patients with conditions such as Parkinson’s disease, stroke, or head and neck cancer \cite{seo2021prevalence,rivelsrud2023prevalence}. It is estimated that around {\$}4.3 to {\$}7.1 billion go towards hospitalization costs associated with dysphagia annually \cite{patel2018economic}. Currently, swallow evaluations are performed through video-based modalities such as flexible endoscopic evaluation of swallowing (FEES), videofluoroscopic swallowing studies (VFSS) or high-resolution manometry, which, while informative, are either invasive, require radiation exposure, necessitate a trained practitioner, or are limited in cost efficiency \cite{video-survey,survey-1}. Given the limitations of current gold-standard swallow evaluations, hospitalized patients at risk for aspiration are typically screened using a clinical swallow evaluation. However, these assessments are performed without instrumentation, demonstrate limited diagnostic sensitivity and specificity, and often leave providers with uncertainty in clinical decision-making. This highlights the pressing need for an objective swallow assessment tool capable of efficiently and accurately identifying dysphagia risk.

To address this gap, we collected audiometric data during fiberoptic endoscopic evaluation of swallowing (FEES) to investigate associations between swallow dysfunction and acoustic features. We then used these data to develop a machine learning model for predicting swallow dysfunction. We present a system that achieved promising classification performance, suggesting that surface digital auscultation captures acoustic signatures with diagnostic value reflective of underlying oropharyngeal dysfunction. 

\vspace{-1mm}
\section{Related Work}
\vspace{-2mm}
To address these challenges imposed by dysphagia assessments, a growing body of research has also explored the use of machine learning techniques to analyze swallow mechanisms and detect related abnormalities using a mix of accelometry, acoustic signals, and electromyography (EMG) \cite{prev-work-0,prev-work-1,prev-work-2,prev-work-3}. While preliminary studies suggest the potential of machine learning for detecting swallow dysfunction, they remain significantly limited. First, and most notably none have conducted real-time assessments of swallow sounds during a gold-standard evaluation such as FEES, restricting their ability to characterize the severity and nature of dysfunction. Second, many studies validated their models using only a single bolus consistency (e.g., a sip of water) \cite{prev-work-0,prev-work-1,prev-work-2,prev-work-3}, despite the fact that dysphagia and aspiration may manifest differently across consistencies- thereby limiting external validity. Finally, most prior work has been constrained to proof-of-concept designs with small sample sizes \cite{smartphone_swallow_detect}. To overcome these limitations, we developed a machine learning model trained on more than 600 swallow events, using acoustic data collected during comprehensive swallow evaluations. 

\vspace{-3mm}
\section{Method}
\label{sec:format}
\vspace{-3mm}

\subsection{Preliminary Experiments}
\vspace{-1mm}

Initial preliminary experiments were conducted to explore the performance of established baseline audio embedding models: AST, CLAP and OPERA \cite{ast,clap,opera} (Figure \ref{fig:overview} section D.1.). This was done with our existing dataset, both with and without demographic features across 5 randomized train–test splits.
``Sev." and ``Abn." standing for severity (3-class) and abnormality (2-class) respectively. 
It is observed that the OPERA model achieve the best performance, hence it was included in our main experiments as a baseline comparison.

Although age and gender did not have substantial improvement to the model's performance, we decided to include it due to prior research indicating that age and gender may change the acoustic characteristics of the swallow\cite{Youmans_Stierwalt_2011}. Furthermore, the performance between RFC and SVM was comparable, but for consistency we chose to use RFC for our main evaluation.

\begin{table}[ht]
\vspace{-2mm}
\centering
\caption{Background Distribution of Participants}
\label{tab:background_distribution}
\vspace{-3mm}
\resizebox{0.9\columnwidth}{!}{ 
\begin{tabular}{lcc}
\hline
\hlineB{2}
\textbf{Factors} & \textbf{Sample Size (n)} & \textbf{Percentage (\%)} \\
\hline
\textbf{Gender} & & \\
\hline
~~Female & 25 & 51.02 \\
~~Male & 24 & 48.98 \\
\hline
\textbf{PAS} & & \\
\hline
~~Normal && \\
~~~~~PAS[1-2] & 22 & 44.90 \\
\hline
~~Abnormal && \\
~~~~~PAS[3-5] & 20 & 40.82 \\
~~~~~PAS[6-8] & 7 & 14.29 \\
~~~~~Total & 27 & 55.10 \\
\hline
\textbf{BMI} & & \\
\hline
~~$<18.5$ (underweight) & 2 & 4.08 \\
~~$\leq24.9$ (normal) & 17 &  34.69\\
~~$>24.9$ (overweight) & 30 & 61.22\\
\hline
\textbf{Age} & & \\
\hline
~~$30$\textendash $60$ & 13 & 26.53 \\
~~$62$\textendash $69$ & 16 & 32.65 \\
~~$70$\textendash $96$ & 20 & 0.41 \\
\hline
\textbf{Chronic Obstructive Pulmonary Disease (COPD)} & & \\
\hline
~~No & 41 & 83.67\\
~~Yes & 8 & 16.33 \\
\hline
\textbf{History of Pneumonia caused by Aspiration (PNA)} & & \\
\hline
~~No & 38 & 77.55 \\
~~Yes & 11 & 22.45 \\
\hline
\hlineB{2}
\end{tabular}
} 
\vspace{-4mm}
\end{table}

\vspace{-3mm}
\subsection{Data Annotation}
\vspace{-2mm}
This study was approved by University of California San Diego, where we recruited a total of 49 participants from Center for Airway, Voice and Swallowing, who self-reported symptoms of dysphagia. These participants underwent swallow evaluations using FEES. A standard FEES includes 8-10 trials of oral intake with different consistencies and bolus sizes. During this, a flexible
laryngoscope is placed through the nose and each swallow trial is observed. Speech and language pathologists and fellowship-trained laryngologists rated each FEES using the penetration-aspiration scale (PAS) to assess the severity of swallow dysfunction. The PAS is a categorical assessment rating, ranging from 1–8, assessing the degree of dysfunction and evidence of oral
intake invading the airway with higher scores indicating higher levels of swallow dysfunction. Based on this scale, PAS of 1-2 are
relatively normal swallows, PAS of 3-5 indicate evidence of penetration, and PAS of 6 and above refers to evidence of aspiration. During the FEES, each participant also had a digital stethoscope (3M TM Littmann Core Digital Stethoscope) placed lateral to the thyroid cartilage to collect audiometric data in real time during the FEES.

In total, 392 audio recordings were collected for this study. Corrupt audio files that contained talking or non-swallow sounds that our parameters could not filter out were removed. The swallows in each recording were then cleaned in preparation for feature extraction with an average duration of 0.64 seconds. After processing, there were 617 individual swallow events (24 participants contributed to 10-15
swallow events each, 10 participants contributed to 15-20 events each, 8 participants contributed $\leq$10 events each and 3 participants contributed $\geq$20 events each).
We aimed to use our domain-informed features compared with features extracted from OpenSMILE and embeddings from OPERA to differentiate patients with swallow dysfunction from those without and possibly reflect the severity of the patients’ condition. The distribution of patients’ background data is shown in Table 1. 

We used the Python Librosa \cite{mcfee_librosa_2025} to process and ``clean" the audio recordings collected for this study and used several parameters to properly identify the start and end (a single segment) of all swallows in each audio file. The parameters consist of 4 elements: (i) A threshold of amplitudes with unit in dB to determine the silence period; 
(ii) The gap time in seconds for which segments that are within this range are merged into a single segment; (iii) The minimum amplitude for which segments with amplitudes less than this value are discarded; and (iv) A maximum amplitude for which segments with amplitudes greater than this value are discarded. By cross referencing with the visual data of the swallows, we tuned the parameters for each audio file individually such that all swallows were segmented correctly 
(Figure \ref{fig:overview} section B.1.).


\vspace{-3mm}
\subsection{Feature Extraction and Modeling}
\vspace{-2mm}
In addition to extracting OPERA embeddings, we also analyzed a series of domain-informed features that are heuristically associated with swallow dysfunction \cite{yousef2025sounding}.

\vspace{-3mm}
\subsubsection{Frequency}
\vspace{-2mm}
To calculate the maximum five frequencies of each audio file, we used 
Fast Fourier Transform. This algorithm efficiently implements the discrete Fourier Transform (DFT) equation, which transforms signals from a time domain \(x[n]\) into its frequency domain \(X[k]\).
DFT Equation: 
\begin{equation} 
X[k] = \sum_{n=0}^{N-1} x[n] \, e^{-j 2\pi k n /N} \end{equation}
Here, \(x[n] \) is the input signal at time index \(n\), \(j\) is the imaginary unit, \(k\) is the index of the frequency bin, \(N\) is the total number of samples and {\(X[k]\)} is the transformed signal in the frequency domain.

Then we used Librosa’s Short-Time Fourier Transform (STFT) function, which repeatedly applies this formula to overlapping windows, to get the entire frequency content of the audio file. Given this frequency content over time, 
 we calculated the average and median frequency over time for each audio file. Finally, we extracted the maximum five frequency values: \(freq_i\) where \(i = 1,..., 5\).

\vspace{-3mm}
\subsubsection{Amplitude} 
\vspace{-2mm}
To get the amplitude- related values, we used Librosa’s load function which converts the .WAV audio files into a time series represented as a 1-dimensional array of floating point values. These values are the respective amplitudes of all samples of the file. Hence, using this array, we calculated the peak (maximum) and average amplitude over time. 

\vspace{-3mm}
\subsubsection{Area Under the Curve} 
\vspace{-3mm}
To calculate the area under the curve, we integrated the absolute waveform using the composite trapezoidal rule provided by the Numpy library. The explicit formula of the composite trapezoidal rule is: 
\begin{equation}
\int_{a}^{b} f(x)\,dx \approx \frac{1}{2} \sum_{j=1}^{n} (x_j - x_{j-1}) \left[ f(x_{j-1}) + f(x_j) \right]
\end{equation}
where \(a, b\) is the interval of integration and \(n\) is the number of partitions.

\vspace{-3mm}
\subsubsection{Representations from Pretrained Audio Models.}
\vspace{-2mm}
Finally, to stay consistent with literature, we also collected all features available from OpenSMILE's \cite{opensmile} toolkit. (For all three methods of extraction, participants’ age and gender was included.) 

\vspace{-3mm}
\subsection{Segmentation and Aggregation Analysis}
\vspace{-2mm}
In clinical settings, audio recordings often contain multiple
swallows within a single trial. Hence, the model is required to detect and classify individual swallows from continuous input. To simulate this, we further evaluate model performance by using both \textit{fixed-parameter} and \textit{sliding window automatic segmentation.} These
methods simulate automated preprocessing in real-world deployment. After generating per-swallow predictions, we
computed patient-level classifications by aggregating results
across all swallows for that patient using three strategies:
Mean-risk (average prediction), Max-risk (highest risk across
swallows), and Mode-risk (most frequent prediction).

\vspace{-3mm}
\subsubsection{Fixed Parameter Segmentation}
\vspace{-2mm}
Fixed parameters were selected by testing threshold combinations and maximising intersection over union (IoU) with ground-truth masks. The best overlap (IoU = 0.4775) was obtained with top\_db = 20, gap\_time = 0.6, allowed\_max\_amplitude = 2, and allowed\_min\_amplitude = 0, yielding 65.8{\%} sensitivity and 87.6{\%} specificity. An example segmentation is shown in Figure \ref{fig:overview} section C.1. with ground truth on top of the predicted segmentation. 
 

\vspace{-1mm}
\subsubsection{Sliding Window Segmentation}
\vspace{-1mm}
The sliding window Segments the audio into 1-second windows with 50\% overlap, ensuring full coverage of the audio timeline. This strategy minimizes the risk of missing swallows and is also algorithmically simple, making it suitable for real-time use. 

\vspace{-1mm}
\subsection{Evaluation Protocol}
\vspace{-1mm}
All experiments were carried out using train-test splits at the patient level to better reflect clinical
use scenarios. This ensures that the model does not overfit to
patients already seen during training. This emulates
our target clinical use case, such that the model is expected
to validate and make inferences on entirely unseen patients.
We performed five independent patient-level splits, with patient ID used as the primary unit of data partitioning. Each split was stratified to
maintain class and swallow distribution.
%
We report AUC-ROC scores as
the primary metric for comparing model performance.

\vspace{-3mm}
\section{Results} \label{sec:pagestyle}  
\vspace{-2mm}

\subsection{Feature Performance Comparison} 
\vspace{-2mm}

\begin{table}[t]  
\caption{Main Results (Patient-Level Splits). ``Sev.'' and ``Abn.'' stand for severity and abnormality respectively.} 
\vspace{-3mm}
\centering  
\setlength{\tabcolsep}{3.5pt}  
\renewcommand{\arraystretch}{1.05}  
\footnotesize
\begin{adjustbox}{max width=\columnwidth}
\begin{tabular}{l|l|ccc}  
\hline  
\hlineB{2}
Task & Method & \makecell{AUC\\ROC} & \makecell{AUC\\PRC} & \makecell{Balanced\\Accuracy} \\ \hline  

Sev. & OPERA     & 0.557 $\pm$ 0.159 & 0.434 $\pm$ 0.130 & 0.542 $\pm$ 0.047 \\ 
Sev. & OpenSMILE (OpSL) & \underline{0.583 $\pm$ 0.120} & \underline{0.503 $\pm$ 0.145} & 0.606 $\pm$ 0.079 \\ 
Sev. & Domain-Informed              & \bf 0.611 $\pm$ 0.055 & \bf 0.519 $\pm$ 0.061 & \bf 0.659 $\pm$ 0.028 \\ 
Sev. & \makecell{Domain-Informed w/ OpSL} & \underline{0.561 $\pm$ 0.135} & 0.493 $\pm$ 0.120 & \underline{0.610 $\pm$ 0.080} \\ \hline  

Abn. & OPERA     & 0.651 $\pm$ 0.176 & 0.718 $\pm$ 0.140 & 0.579 $\pm$ 0.080 \\ 
Abn. & OpenSMILE & 0.778 $\pm$ 0.144 & 0.850 $\pm$ 0.094 & 0.665 $\pm$ 0.152 \\ 
Abn. & Domain-Informed              & \textbf{0.904 $\pm$ 0.015} & \bf 0.913 $\pm$ 0.075 & \bf 0.755 $\pm$ 0.061 \\ 
Abn. & \makecell{Domain-Informed w/ OpSL} & \underline{0.804 $\pm$ 0.183} & \underline{0.862 $\pm$ 0.081} & \underline{0.710 $\pm$ 0.159} \\ 
\hline
\hlineB{2}

\end{tabular}  
\end{adjustbox}
\label{main}  
\vspace{-2mm}
\end{table}  

The 2-class (abnormality detection) model trained with our domain-informed features performed the best with an \textbf{AUC-ROC of 0.904} compared to the pretrained audio embeddings (Table \ref{main}).
Combining OpenSMILE with our domain-informed features decreases the model performance to a AUC-ROC score of 0.804 in a binary-class setting. This makes sense as the acoustic features of OpenSMILE may capture irrelevant or noisy characteristics.
Overall performance in the 3-class (severity classification) setting is notably lower than in the binary-classification setting. This is largely due to the smaller training samples per-class which hinders the model's ability to learn discriminative patterns for each category. Further exploration with larger and more balanced datasets may help unlock the potential of multiclass swallowing classification.

\vspace{-3mm}
\subsection{Segmentation and Patient Aggregation Results}
\vspace{-2mm}

\begin{table}[t]
\centering
\caption{Evaluation with Audio Segmentation, Aggregated by Patient (AUC-ROC Scores)}
\vspace{-3mm}
\label{tab:audioseg}
\setlength{\tabcolsep}{3.5pt}
\renewcommand{\arraystretch}{1.05}
\begin{adjustbox}{max width=\columnwidth}
\begin{tabular}{l|ccc}
\hline
\hlineB{2}
\textbf{Method} & \textbf{Mean} & \textbf{Max} & \textbf{Mode} \\ \hline
Sliding Window           & \underline{$0.893 \pm 0.103$} & $0.856 \pm 0.106$ & \underline{$0.884 \pm 0.104$} \\ 
Fixed-Parameters         & $0.868 \pm 0.142$ & $\mathbf{0.942} \pm 0.051$ & $0.842 \pm 0.141$ \\ 
Human Segmented Swallows & $\mathbf{0.967} \pm 0.054$ & \underline{$0.918 \pm 0.079$} & $\mathbf{0.971} \pm 0.041$ \\ 
\hlineB{2}
\end{tabular}
\end{adjustbox}
\vspace{-4mm}
\end{table}

The Human Segmented Swallows condition serves as a benchmark, representing the scenario where all swallows are accurately segmented (Table \ref{tab:audioseg}). The Fixed parameter segmentation method produces promising results, with the highest AUC-ROC score of 0.942 observed under the Max-risk aggregation strategy. However, we note that under Mean-risk and Mode-risk aggregation, the Sliding-window segmentation outperforms the fixed-parameter approach. This may suggest that while the fixed-parameter \& max-risk result is strong, it may be a dataset-specific artifact rather than a consistently superior configuration. 


The confusion matrices uses Human Segmented Swallows, illustrating that the model is effective in identifying at-risk patients with minimal false negatives (Figure \ref{fig:overview} section D.2). This is clinically significant, as misclassifying high-risk patients could result in missed patients with unsafe swallows being cleared for an oral diet putting them at risk for aspiration
pneumonia.

\vspace{-1mm}
\subsection{Feature importance and statistical analysis}
\vspace{-1mm}

\begin{figure}[ht]
\vspace{-4mm}
\centerline{\includegraphics[width=0.8\columnwidth]{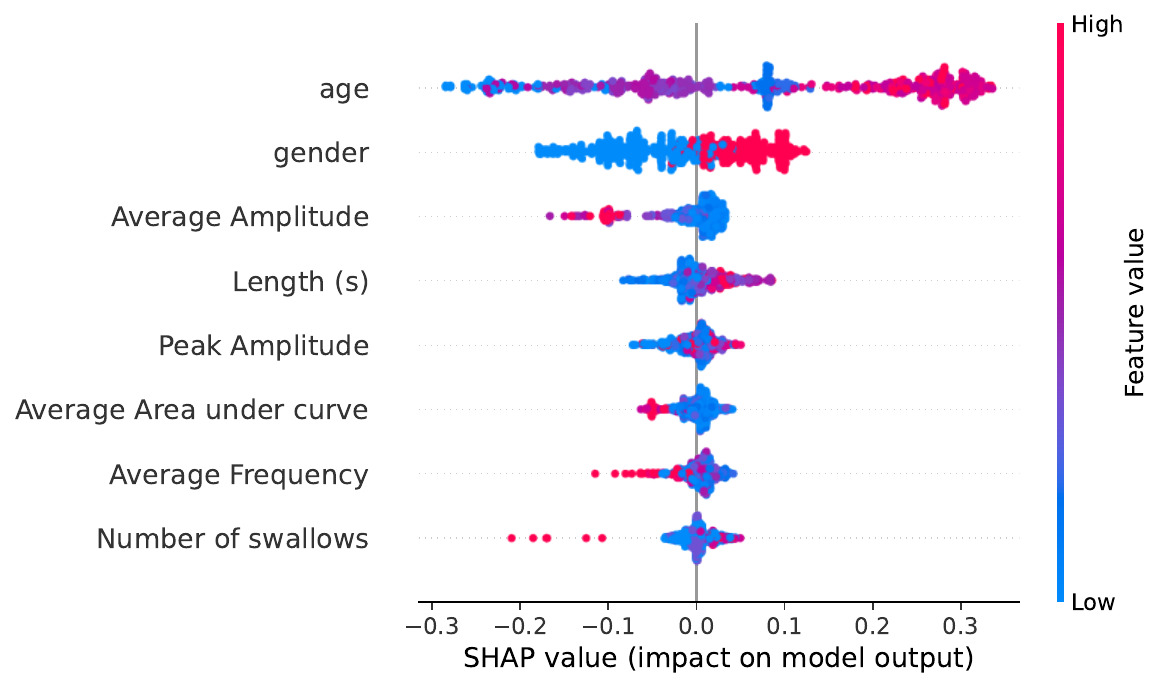}}
\vspace{-3mm}
\caption{SHAP Summary Plot of Top 8 Features from Performance on Human Segmented Swallows}
\label{fig:shap}
\vspace{-2mm}
\end{figure}

SHAP analysis (Fig.\ref{fig:shap}) shows that age and gender are influencial predictors of dysphagia, with older age and male sex associated with higher risk, consistent with prior studies.\cite{Bolinger2020, Tomonaga2024} Signal-based features, average amplitude, frequency, area under the curve, and swallow count also strongly influenced classification, with weaker swallows linked to dysphagia. These findings support integrating both demographics and acoustic characteristics maximizes predictive accuracy.

%
%
Lastly, we compared the accuracy of using our domain-informed features to previously created acoustic datasets (OPERA \& OpenSMILE) and found that our domain-informed method outperformed all other approaches, as shown by the critical difference test (Figure \ref{fig:overview} section D.3). 

\vspace{-2mm}
\section{Conclusion}
\vspace{-2mm}
This study presents a machine learning framework for detecting dysphagia using noninvasive acoustic sensing. By capturing neck-region acoustic signals and leveraging time-frequency analysis, our system demonstrates the potential for identifying abnormal physiological patterns of swallowing with high accuracy. The proposed approach offers a low-cost, scalable alternative to traditional diagnostic methods and represents a step toward accessible, real-time pharyngeal health monitoring.

\textbf{Limitation and Future Work.}
While promising, 
the dataset size remains moderate, limiting generalization across diverse populations and conditions. Expanding the dataset, improving the automatic swallow segmentation algorithm and validating in diverse scenario, such as at-home settings, will be essential next steps.

\clearpage
\section{Acknowledgments}
We thank medical students Sacha Moufarrej, Bao Luu, and David Zeng from the University of California San Diego School of Medicine, for their assistance with splicing the swallow audio files. No external funding was received for this study. The authors declare no relevant financial or nonfinancial conflicts of interest. All data were collected from patients as part of an approved study and were fully anonymized prior to analysis.

\bibliographystyle{IEEEbib}
\bibliography{strings,refs}

\end{document}